\documentclass{article}
\usepackage[a4paper,top=3cm,bottom=2cm,left=3cm,right=3cm,marginparwidth=1.75cm]{geometry}

\usepackage[utf8]{inputenc} 
\usepackage[T1]{fontenc}    
\usepackage{hyperref}       
\usepackage{url}            
\usepackage{booktabs}       
\usepackage{amsfonts}       
\usepackage{nicefrac}       
\usepackage{microtype}      
\usepackage{amssymb}
\usepackage{amsmath}
\usepackage{mathtools,mathrsfs}
\usepackage{blkarray}

\usepackage{comment}
\usepackage{xcolor}
\title{Fitting Elephants}

\author{
Partha P Mitra  \\
  Cold Spring Harbor Laboratory, Cold Spring Harbor\\
  NY, NY 11724\\
  Center for Computational Brain Research, IIT Madras\\
  Chennai, India\\
  \texttt{mitra@cshl.edu} \\
}

\begin{document}

\maketitle

\begin{abstract}
{\color{black} Textbook wisdom advocates for smooth function fits and implies that interpolation of noisy data should lead to poor generalization. A related heuristic is that fitting parameters should be fewer than measurements ({\it Occam’s Razor}). Surprisingly, contemporary machine learning (ML) approaches, {\it cf.} deep nets (DNNs), generalize well despite interpolating noisy data. This may be understood via {\it Statistically Consistent Interpolation} (SCI), {\it i.e.} data interpolation techniques that generalize optimally {\it for big data}. \textcolor{black}{In this article we elucidate SCI using the weighted interpolating nearest neighbors ({\it wiNN}) algorithm, which adds singular weight functions to {\it kNN} (k-nearest neighbors). This shows that data interpolation can be a valid ML strategy for big data.} SCI clarifies the relation between two ways of modeling natural phenomena: the rationalist approach (strong priors) of theoretical physics with few parameters and the empiricist (weak priors) approach of modern ML with more parameters than data. SCI shows that the purely empirical approach can successfully predict. However data interpolation does not provide theoretical insights, and the training data requirements may be prohibitive. Complex animal brains are between these extremes, with many parameters, but modest training data, and with prior structure encoded in species-specific mesoscale circuitry. Thus, modern ML provides a distinct epistemological approach different {\it both} from physical theories and animal brains.}


\end{abstract}


{\bf Introduction:} We can easily imagine the perfect equality of two sticks, even though we have never seen two real sticks being exactly equal. Thus, argues Socrates in Phaedo, we must have been born with the {\it idea} of perfect equality, and it was not derived from our sensory experience. The existence and extent of {\it a priori} knowledge has long been a subject of philosophical debate between the rationalist and empiricist schools of thought, the latter emphasizing the derivation of all knowledge from sense data. This debate has re-emerged in the enterprise of modern machine learning (ML), with the failures of a previous generation of AI attributed to strong and possibly inappropriate priors in the form of hand-crafted rules. In contrast, the idea is that the modern approach succeeds by focusing on empirical learning from exhaustively large data sets without incorporating strong priors. 

\textcolor{black}{Two striking and inter-related features of current practice in ML are over-parameterization (more model parameters than data values) and data interpolation (exactly fitting the training data labels). Traditionally, in statistics, the practice has been to minimize the number of model parameters and not to interpolate the training data, as this is thought to lead to fitting "noise" rather than "signal" and consequently to poor generalization on a test data set\cite{james2013introduction,gyorfi02}. Over-parameterization also runs counter to scientific theoretical norms of model parsimony for understanding natural phenomena, as exemplified by theoretical physics. Another important context comes from neurobiology and the nature-nurture tradeoff. Although shaped by experience and learning, brain architecture is encoded in the species genome and unfolded through a developmental program, a modern day version of Socrates' prior knowledge. Thus the emerging practice of interpolating noisy data distinguishes modern ML from textbook statistics, theorizing in the physical sciences as well as biological learning with its species-specific priors.}

\textcolor{black}{Given this departure from standard practice in multiple disciplines as well as from biological learning, it is particularly striking to note that commercially successful modern ML approaches generalize well even with training data interpolation\cite{zhang2017understanding}. The circumstances under which this can happen is the subject of a growing body of research\cite{zhang2017understanding,wyner2017explaining,belkin2018understand,belkin2018overfitting,cutler2001pert}. The current article attempts to clarify one of the key underlying issues, namely the {\it statistical consistency of data interpolation}, {\it i.e.} the circumstances under which the data interpolation function is {\it optimal} in the sense of having the least possible generalization error in the limit of large data sizes.}

\textcolor{black}{This phenomenon, Statistically Consistent Interpolation or SCI, may be readily understood in terms of a newly introduced class of non-parametric interpolating function estimators discussed in the article, and sheds light on the differing approaches to modeling natural phenomena provided by traditional physical theory, modern ML, and biological brains. Beyond the theoretical interest, SCI has practical implications. It opens up a new area of research by bringing function interpolation techniques, an established discipline but applied previously largely to noise free data (such as in computer graphics), to bear on ML problems. It also clarifies the circumstances under which data interpolation using modern ML approaches might be legitimate in scientific data analysis despite the presence of measurement noise.}

\begin{figure}
  \centering
    \includegraphics[width=\linewidth]{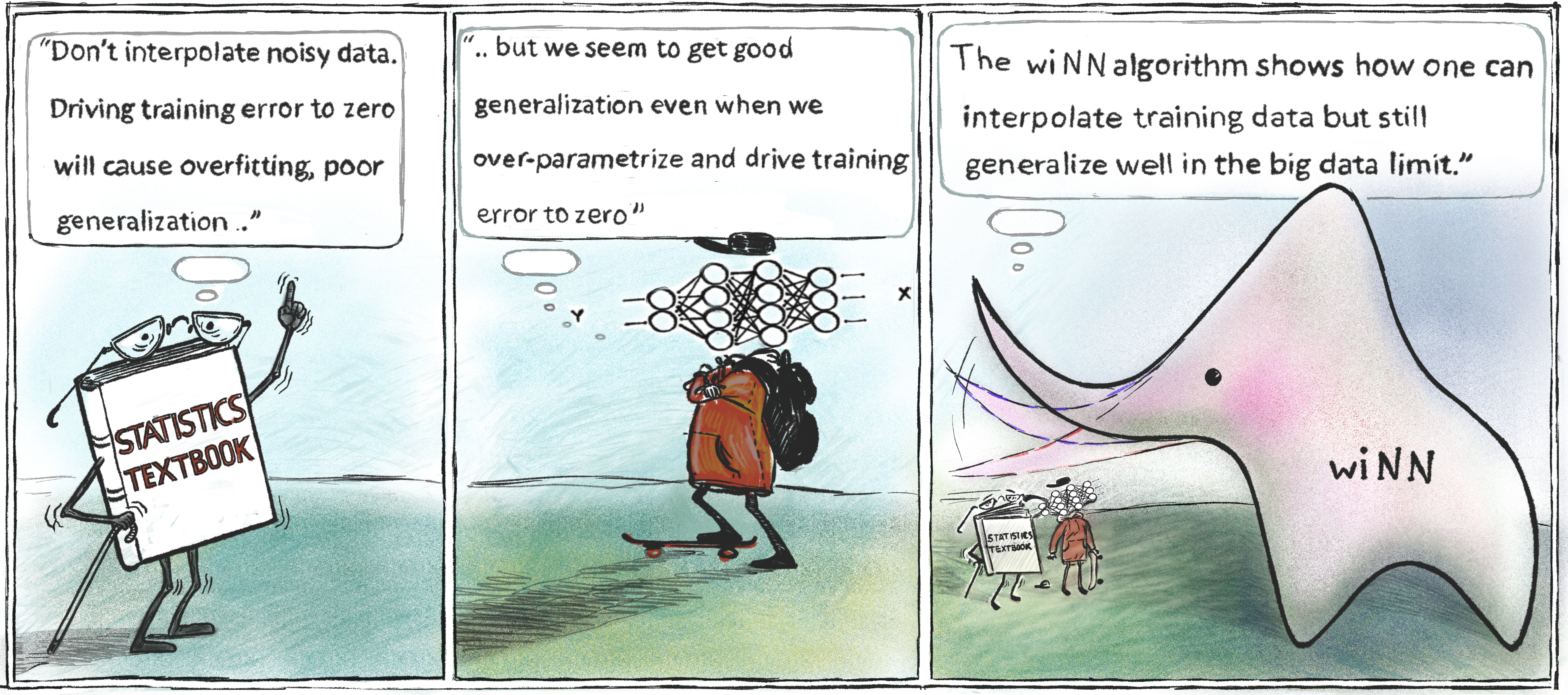}
\end{figure}

{\bf Disciplinary context: physical sciences, classical statistics and computer science} Theoretical physics has long sought to minimize the number of free parameters, as exemplified by Von Neumann's quip\cite{dyson2004meeting} about being able to "fit an elephant" given four parameters, and making its tail wiggle if given a fifth. An important triumph of theoretical physics has been to encapsulate natural phenomena in a small number of laws with a very small number of fundamental parameters. Some of the most important aspects of physical law (e.g., the conservation of mass-energy) do not involve any parameters whatsoever, and do not relate in any obvious way to parametric model fitting, \textcolor{black}{let alone data interpolation}. 

Classical statistics also aims to minimize fitting parameters and in addition cautions against \textcolor{black}{data interpolation} (e.g., see P.21 \cite{gyorfi02}). In modern statistics, algorithms such as $l_1$ penalized sparse regression\cite{DonohoPT} have explored the high-dimensional regime\cite{wainwright2019high} where the number of fitting parameters and data points are comparable. However, standard statistical practice is to use  non-zero regularization to maintain non-zero error on the training set and not to interpolate data. Indeed, it has been suggested that a reason why data interpolation has taken hold in modern machine learning is that computer scientists are not used to making strong distinctions between signal and noise\cite{wyner2017explaining} and are therefore more comfortable "fitting noise". 

In light of these divergent views and practices across disciplines, the pragmatic success of noisy data interpolation using overparameterized learners \cite{belkin2018understand} for real applications, calls for better theoretical understanding. \textcolor{black}{The success of interpolating learners also has important practical implications - it brings a rich set of theoretical and computational tools for data interpolation, previously used in noise-free settings, into the arena of machine learning, and has the potential of changing how data analysis is taught and practiced.}

\begin{figure}
  \centering
    \includegraphics[width=\linewidth]{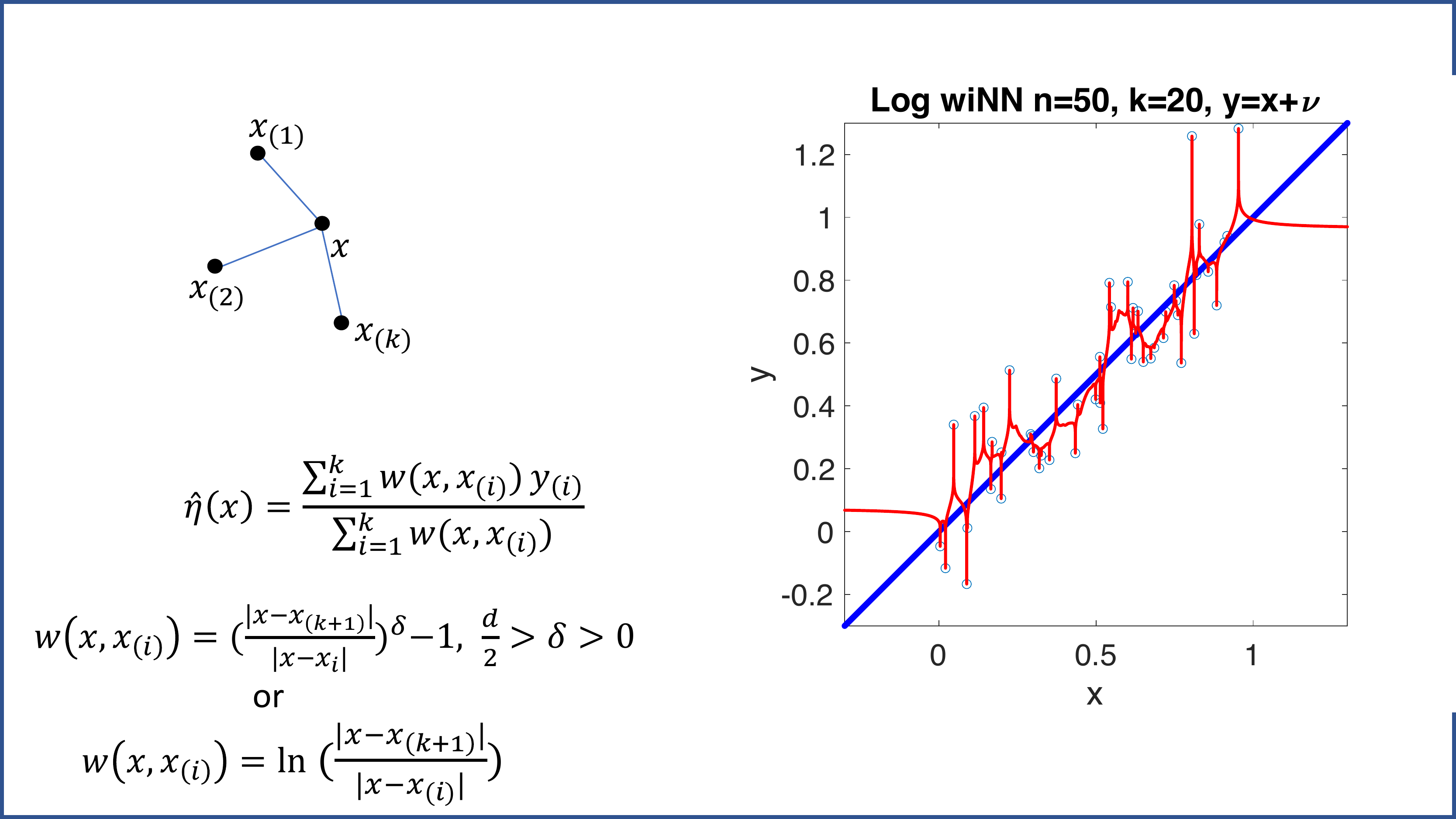}
   
    \caption{The Weighted Interpolating  Nearest Neighbor (wiNN) algorithm and its application to a simple linear regression problem. In the regression example, the sample data points (open circles) are given by $y_i=x_i+\nu_i$ ($i=1..50$) where $x_i$ are uniformly distributed over the interval $[0~1]$ and the noise is normally distributed $\nu_i\sim N(0,1)$. $y=x$ is given as a solid blue line. The wiNN interpolator using logarithmic weights and $k=20$ is shown as a red curve (note that the curve is close to the blue line, except near the data points where it spikes to interpolate). Outside the range of $x_i$, the interpolating function goes to a constant value consistent with its value at the edge of the range: wiNN is an interpolation algorithm, and cannot extrapolate.}
\end{figure}

{\bf How can noisy data interpolation coexist with good generalization?} 

The literature on overparameterized and interpolating learners is growing rapidly and is beyond the scope of this perspective to review\cite{zhang2017understanding,wyner2017explaining,rakhlin2018consistency,ongie2019function,belkin2019reconciling,liang2018just,bartlett2019benign,montanari2019generalization,karzand2019active,xing2018statistical}. However, it is useful to briefly discuss some of the relevant ideas. 

Much of the current theoretical analysis in machine learning is centered around the computation of bounds on the generalization error based on the capacity of the class of fitting functions(\cite{ab-nnltf-99}, \textcolor{black}{also see Rademacher complexity analysis of overparameterized  networks\cite{arora2019fine,allen2019learning}).} However, model complexity based bounds lack explanatory power for interpolating learners even for the theoretically well controlled case of Kernel Machines\cite{belkin2018understand}. One set of theoretical ideas to explain good generalization under zero training error conditions can  be found in Boosting, where weak learners are progressively added while re-weighting training examples to prioritize difficult cases. Continuing to boost can drive the training error for classification to zero, without increasing the test error (see Fig.1 in \cite{schapire1998boosting}), a phenomenon that has been explained by invoking the concept of classifier margin on the training set. However, the margin theory of boosting does not directly apply to regression functions that interpolate data labels since the margin on the training set is zero.

Deep nets are non-linearly parameterized functions, and are challenging to analyze theoretically except in limiting cases. \textcolor{black}{In addition, an interpolating deep net fit does not guarantee good generalization \cite{mucke2019global}. The phenomenon of SCI, although observed empirically while training deep nets, is better understood directly within the context of a simple non-parametric estimation algorithm presented below.} We will comment \textcolor{black}{briefly} on parametric models such as deep nets in a later section. 

\underline{Statistically consistent interpolation: the {\it wiNN estimator}} 

A recently proposed non-parametric data interpolation algorithm for classification and regression, the (singularly) weighted Interpolating Nearest Neighbors algorithm ({\it wiNN})\cite{belkin2018overfitting} (Figs.1-2) helps understand why zero training error can coexist with good generalization\cite{belkin2018overfitting} for broad function classes. The {\it wiNN} algorithm belongs to the class of Nadaraya-Watson kernel estimators\cite{nadaraya1964estimating,watson1964smooth} but utilizes a {\it singular} kernel. Here we briefly review this algorithm together with proof sketches (provided in supplementary section 1), illustrative examples and a contextual discussion. 

\textcolor{black}{Consider $n$ labelled data points $x_i,y_i$ drawn from a suitable joint distribution $P(x,y)$, where $(x_i,y_i)\in R^d\times R$ for regression and $(x_i,y_i)\in R^d\times \{0,1\}$ for classification. The {\it wiNN} estimator (with parameters $k,\delta$) for the regression function ({\it i.e.} conditional mean) at a test point $x$ is given by a singularly weighted average of the $k$ nearest neighboring sample points $x_{(i)}$, $i=1..k$, where $x_{(i)}$ is the $i^{th}$ closest neighbor of $x$ (Fig.1): }
\begin{equation}
    \hat{\eta}(x)=\frac{\sum_{i=1}^k y_{(i)} w(x,x_{(i)})}{\sum_{i=1}^k w(x,x_{(i)})}
    \label{wiNNeqn}
\end{equation}
where the weights $w(x,x_{(i)})\coloneqq (r_{(k+1)}/r_{(i)})^\delta-1$ with $0\leq \delta < d/2$. The $\delta=0$ case corresponds to logarithmic weights $w(x,x_{(i)})\coloneqq log(r_{(k+1)}/r_{(i)})$. Here $r_{(i)}=|x-x_{(i)}|$. The weights can be generalized as long as the singularity at the origin (which enforces training data interpolation) is maintained. Note that if $1$ is not subtracted in the definition, then the $\delta=0$ case corresponds to the classical k-Nearest Neighbor (kNN) estimator. The original proposal in \cite{belkin2018overfitting} does not include the $-1$ in the definition but we find it convenient to include so as to connect to the logarithmic weights. For classification, the plug-in estimator for the label at a test point is obtained by thresholding $\hat{\eta}(x)$. 

The behavior of this estimator is illustrated in Fig.1 for regression and Fig.2 for classification. Note that the estimator is {\it local} and {\it data adaptive} (through the choice of the neighborhood parameter $k$). Both are crucial properties of the estimator, and are shared with classical $kNN$ style estimators - the novelty is that the  $wiNN$ estimator interpolates.  

\begin{figure}
  \centering
       \includegraphics[width=\linewidth]{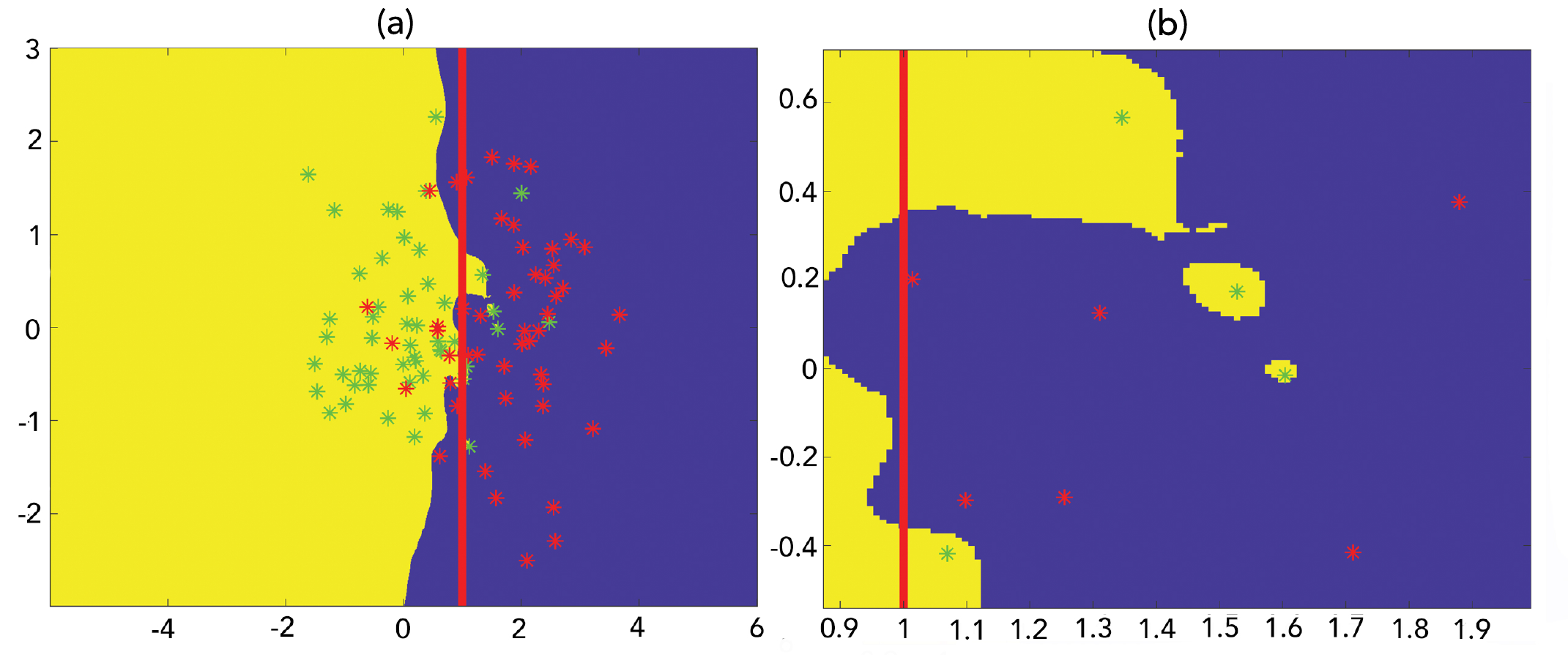} 
 \caption{Classification using wiNN is illustrated in (a) in 2D; $P(y=1|x)\sim N((2,0),1)$ and $P(y=0|x)\sim N((0,0),1)$. $N((2,0),1)$ and $N((0,0),1)$ denote isotropic 2D Gaussian distributions centered at $(2,0)$ and $0,0$, each with unit variance. The sample points in the training set are marked with red and green symbols. The red vertical line is the Bayes classification boundary, and the wiNN plug-in estimators using log weights ($n=50$, $k=20$) for the two classes are shown in purple and yellow. (b) Zoom-in portion of the image on the right shows islands of mis-classification around the mis-labelled green stars (cf: adversarial examples)}
\end{figure}

\underline{\it Precursors:} 

The well-known "one nearest neighbor" ($1NN$) rule is in fact an interpolating classifier. It is worth emphasizing that in the large data limit, this well-known rule has a classification risk that is within a factor of two of the  Bayes risk\cite{cover1967nearest}, $\lim_{n\rightarrow\infty}R_{1NN}\leq 2 R_B$. The $1NN$ rule is {\it not} statistically consistent (i.e. the asymptotic large-sample limit is not equal to the Bayes risk). However, if label noise is small, this is already quite good. If the classes do not overlap and there is no label noise, then the $1NN$ rule predicts perfectly in the large sample size limit. For example, if the label noise is low such that only $5\%$ of the samples are mislabelled, then the generalization error for the $1NN$ rule will be less than $10\%$ {\it in the big data limit}. Statistical consistency can be achieved by using the $kNN$ rule, where the labels of $k$ nearest neighbors are averaged over, but the $kNN$ estimator smooths and does not interpolate the training data. 

Two previous examples of using singular weights for interpolation exist: Shepard's method\cite{shepard1968two} for spatial interpolation of geophysical data in $2D$ and $3D$, and the non-adaptive Hilbert-kernel estimator due to Devroye {\it et al}\cite{devroye1998hilbert}, with $k=n$ and $\delta=d$, however, in general, interpolating learners for noisy data are missing from classical statistics. 

An interesting side note:  the well-known Lagrange polynomial interpolation\cite{waring1779vii}  in 1D can be recovered from Eq.\ref{wiNNeqn} with the singular "weight" $w(x,x_i)=w_i/(x-x_i)$ with  $w_i=\prod_j (x_j-x_i)^{-1}$, but note that this "weight" function can be negative and Lagrange interpolation is not a very good interpolation scheme for general positions of the interpolation points (due to Runge's phenomenon\cite{runge1901empirische}). 

\underline{\it Statistical consistency of {\it wiNN}:} 

\textcolor{black}{Importantly, it can be shown \cite{belkin2018overfitting}(an intuitive proof-sketch is provided in the supplementary section 1), that the {\it wiNN} estimator is {\it statistically consistent} for functions (or class-conditional PDFs, respectively, for regression and binary classification problems) chosen from H{\"o}lder classes, a broad function class.} This means that in the case of classification, the expected risk or generalization error tends to that of the Bayes classifier in the large sample limit $n\rightarrow\infty$. For regression, the expected risk or generalization error is measured using the expected mean squared error between the sample-based estimate of the regression function and the true regression function. These expected risks can be shown to approach the theoretical lower bounds as the number of samples grow. 

More precisely, for regression the excess risk goes to zero as (see the supplementary  section for notation and a proof sketch):
\begin{equation}
R_{sq}(\hat{\eta})-R_{sq}(\eta)=E[(\hat{\eta}(x)-\eta(x))^2]\sim n^{-\frac{2\alpha}{2\alpha+d}}
\end{equation}
For classification, 
\begin{equation}
    E[R_{0/1}(\hat f)]-R_{0/1}(f_B) \leq \sqrt{E[(\hat{\eta}(x)-\eta(x))^2]}\sim n^{-\frac{\alpha}{2\alpha+d}}
\end{equation} 
 
 \textcolor{black}{Here $R_{sq}$ is the expected value of the squared loss or risk, $\hat{\eta}$ is the wiNN estimator, $\eta$ is the regression function, $R_{0/1}$ is the zero-one loss for classification, and $f_B$ the Bayes classifier.} 
 The basic idea is to use a bias-variance decomposition of the excess risk term. The variance term reduces with increasing $k$ as $\sim\frac{1}{k}$, as would be expected from averaging $k$ samples. The bias term increases with $k$ since further neighbors are used in the estimator, however if the sample size $n$ increases for fixed k, then the distance to the $k^{th}$ neighbor also decreases, so that bias decreases with increasing $n$.  Under the assumptions of $\alpha-$Holder continuity of the regression function in dimension d, the bias term behaves as $(\frac{k}{n})^{\frac{2\alpha}{d}}$. Trading off the bias and variance terms, one finds that the optimal sample-size dependent choice for $k$ is $\sim n^{\frac{2\alpha}{2\alpha+d}}$. For this choice of $k$, the excess risk for regression goes to zero as $n^{-\frac{2\alpha}{2\alpha+d}}$. 
 
\textcolor{black}{Not only is the wiNN estimator consistent under general conditions, surprisingly it can {\it outperform} the smoother non-interpolating kNN estimator for suitable choices of parameters\cite{xing2018statistical,xing2019benefit}.} This goes against the grain of the intuition that smoother non-interpolating regression functions are better than interpolating regression functions that appear less smooth.  

Note that {\it wiNN} exhibits the "curse of dimensionality" as one would expect given its general setting. The number of samples needed to reach an excess risk $\sim\epsilon$ scales exponentially with dimension as $n(\epsilon)\sim (1/\epsilon)^{1+\frac{d}{2\alpha}}$. 

\underline{\it How do deep net based function fits relate to SCI?} 

\textcolor{black}{A detailed theoretical treatment of the generalization properties of overparameterized deep networks that interpolate data is beyond the scope of this article, but we present some relevant ideas here. The possible connections between these different interpolation schemes in the context of statistical consistency is schematically illustrated in Fig.3.}

\textcolor{black}{The {\it wiNN} algorithm is a local, non-parametric function estimator and exhibits SCI in any data dimension for a broad class of functions. A fixed size DNN is a parametric function estimator and cannot in general be statistically consistent for a similarly broad class of functions. It will generally stop having the capacity to interpolate as the training data set size grows indefinitely. Furthermore, a DNN with a fixed number of parameters cannot be expected to exhibit the same locality properties. However, if the number of parameters in a DNN are allowed to increase with the sample size in a sufficiently flexible way to permit training data interpolation, together with an appropriate optimization procedure for selecting an interpolating DNN for a given sample size, one would effectively obtain a non-parametric function fitting procedure. It is reasonable to ask whether statistical consistency can be obtained in a suitable limit (such as infinitely wide deep nets).}

\textcolor{black}{Whether such a limiting procedure can lead to statistical consistency in {\it any} input dimension for some appropriate class of DNNs, is not currently known. However, it is conceivable that such a limiting procedure could produce DNN based function interpolators that are statistically consistent in the asymptotic limit of {\it large input dimensionality.} To see why this may be the case, first note the example of Simplex interpolation, an algorithm also presented in \cite{belkin2018overfitting}, which is not statistically consistent in finite dimensions, but where the  generalization gap goes to zero exponentially with the input dimension. The algorithm is conceptually straightforward and is based on constructing a simplicial complex with the input data vectors, and applying linear interpolation inside each simplex. This indicates the existence of a class of interpolators that are {\it asymptotically consistent} with increasing input dimensionality.} 

\textcolor{black}{Secondly, it has been shown that Kernel Machine based interpolation can exhibit asymptotic consistency in high dimensions \cite{rakhlin2018consistency}. Note that Kernel Machine Regression has locality properties ({\it c.f.} locality of Spline interpolation). Finally, Kernel Machine regression can be linked to DNNs in the limit of infinitely wide nets ({\it c.f.} the Neural Tangent Kernel\cite{jacot2018neural}). Thus, it is possible that suitable classes of DNNs could exhibit statistical consistency after taking the appropriate limits. This is an interesting challenge for future research.}

\begin{figure}
  \centering
       \includegraphics[width=\linewidth]{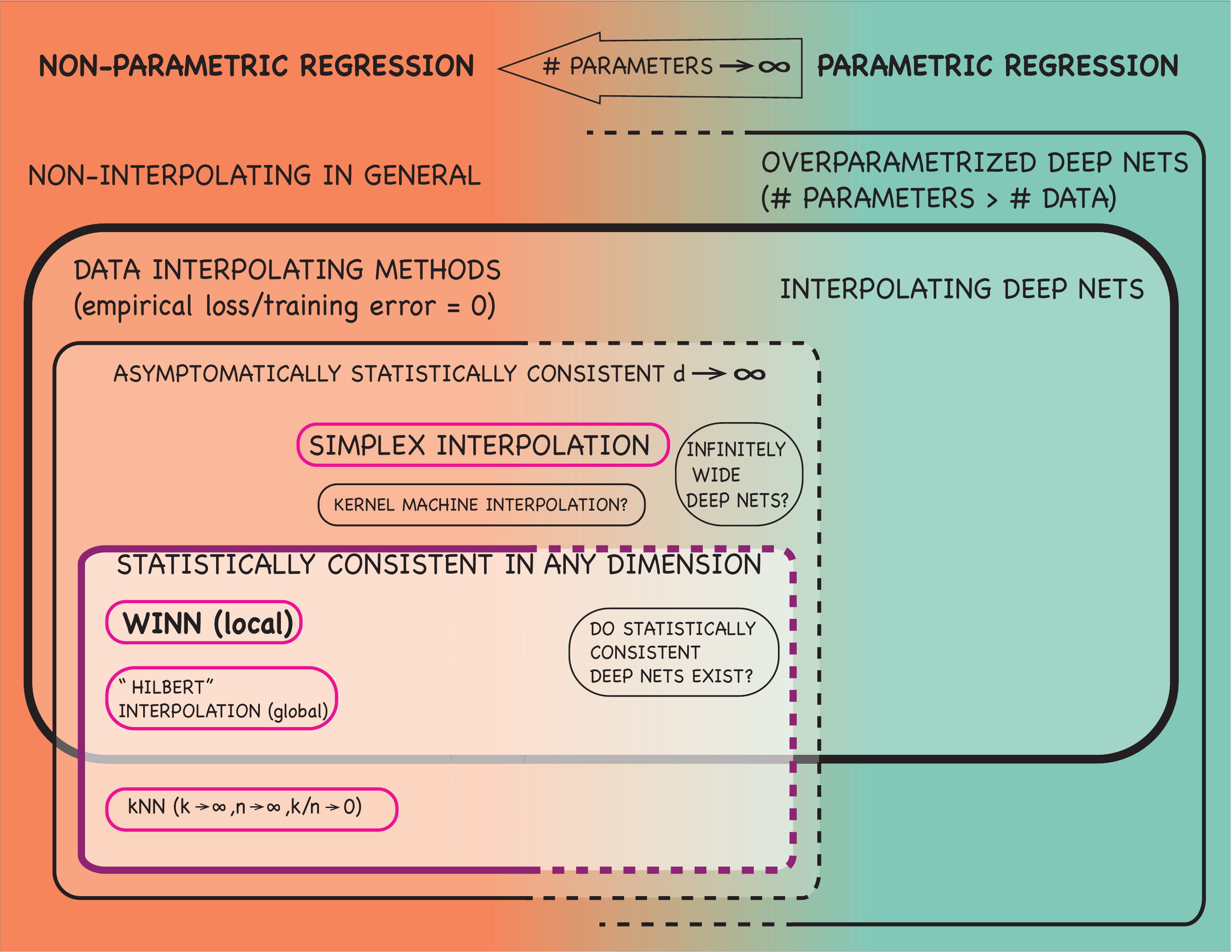} 
    \caption{\textcolor{black}{Tentative landscape outlining the relation between statistically consistent regression estimators, data interpolation, and parametric vs non-parametric estimation. Interpolating estimators may be either non-parametric or parametric. Note that in discussing statistical consistency we have in mind sufficiently general function classes, cf. H{\"o}lder spaces - the discussion here is meant to be suggestive rather than formal. Estimators may be statistically consistent in any input data dimension, or may be asymptotically consistent as the input dimension grows indefinitely. Statistically consistent estimators do not have to interpolate training data (and previous studies have focussed on regularized estimators that do not interpolate), but parametric estimators are not generally expected to be statistically consistent for broad function classes unless suitable limits are taken where they effectively behave like non-parametric estimators. Only two interpolating estimators are currently proven to be consistent for broad function classes in any dimension (wiNN and the Hilbert interpolator). The Simplex interpolation method was proven to be statistically consistent in the limit of large data dimensions. It is not currently known if arbitrary Kernel Machine interpolators are asymptotically consistent, but some theoretical progress has been made in this regard. Infinitely wide nets may resemble Kernel Machine interpolators, but it is not currently known if interpolating deep nets can exhibit statistical consistency for broad function classes in arbitrary dimensions.}}
\end{figure}

To invoke {\it wiNN}-like interpolating algorithms to explain the \textcolor{black}{empirically observed} success of overparameterized deep nets, one needs the additional element that the {\it effective} dimensionality of the input space $d_{eff}$ is small enough that training data set sizes match contemporary big data. If $d_{eff}\sim 10$ and $\alpha=1$, then to achieve $90\%$ accuracy one needs $\sim 10^{1+10/2}=10^6$ data samples - and millions of labelled samples are now routine in many domains. \textcolor{black}{Note that it is not necessary for an interpolation scheme like wiNN to learn a low dimensional representation. Representation learning has been suggested as one of the reasons for the success of DNNs \cite{Goodfellow-et-al-2016}, and may well play a practical role, but the existence of nonparametric interpolation schemes seem to indicate that (parametric) representation learning is not necessary for achieving statistical consistency in the big data limit.}

In essence, what is being said here is that modern overparameterized learning machines ({\it c.f.}, deep nets) resemble classical non-parametric estimators, operating in an interpolating regime and with asymptotically large training data sets. Labelled data sets have gotten so large that even with the curse of dimensionality, local non-parametric estimation can now work for complex input spaces such as natural images. Although the nominal dimensionality of such input spaces would still appear to be too big, natural structure in the real world ({\it e.g.}, the data lies on a low dimensional manifold) would then permit the success of this kind of effectively local, non-parametric estimation. 

A somewhat different approach to the generalization properties of overparameterized ANNs, compared with the "local averaging" picture presented above, is from the perspective of norm-controlled function approximation. It has been argued that good generalization in deep nets is achieved by controlling the norm of the weights rather than the number of parameters (note that such norm control is essential for infinitely large networks)\cite{ongie2019function,bartlett1997valid,neyshabur2014search}. Fitting a single hidden layer ReLU network with a constraint on the Euclidean norm of the weights has been shown to be equivalent to choosing a function while minimizing a specific norm\cite{ongie2019function}. Notably, this norm is {\it not} an RKHS norm and therefore does not fit simply into the above discussion relating large ANNs to GPs. Clearly, there are a number of open theoretical directions to pursue.

\underline{\it Costs and benefits of interpolating learners} 

There are two "costs" of learning via interpolation that also correspond to phenomena that have been observed with deep nets: {\it non-extrapolation}, and {\it adversarial examples}.

First, \textcolor{black}{interpolating learners will not in general extrapolate} (unless, of course, supplemented by prior knowledge about the function outside the data range, which is not a possibility we will consider here). As seen in Fig.1, at the edge of the training data set, the regression function becomes a constant given by its value at the edge. This is consistent with the difficulties encountered by deep nets when going outside the domain where significant amounts of training data are present. This issue raises its head in different guises, such as "continual learning" for non-stationary data, or "transfer learning" across domains, neither of which have fully satisfactory solutions. \textcolor{black}{A recent real life example of the lack of extrapolation, is provided the numerous and widespread glitches that appeared in contemporary AI engines due to a sudden domain shift caused by the Covid pandemic\cite{heaven2020our}.}

Second, the wiNN algorithm provides a ubiquitous mechanism for {\it adversarial examples}\cite{goodfellow2014explaining}, that are well known for deep nets. \textcolor{black}{For example, adversarial examples in image classification correspond to the surprising mis-classification of images due to visually imperceptible addition of small but intentionally adversarial noise\cite{goodfellow2014explaining}, or even rotation/translation of the images after padding\cite{engstrom2017rotation}.} As can be seen in Fig.2, the singular weights forces mis-classification in the immediate vicinity of a mislabeled sample. In regions of the input space where there is some finite density of mislabeled points, adversarial examples are therefore dense \cite{belkin2018overfitting} in the large sample limit (but occupy a set of measure zero since the training set is countable, and therefore adversarial examples must be searched for and cannot be found "at random"). 

There is however an important benefit of overparameterized, interpolating learners, in terms of the learning dynamics. With significant overparameterization, data-interpolating global minima with zero training error proliferate and can be relatively easily found using stochastic gradient descent, in contrast with the under-parameterized case where the optimization dynamics can get frequently stuck in local minima due to the non-convexity of the landscape. If the loss function possesses second derivatives around these global minima, then the asymptotic convergence rates may be analyzed transparently using an effective linear regression model that permits closed form theoretical analyses\cite{ma2017interpolation, mitra2018fast, jacot2018neural}. In general, the interpolating limit appears to provide tractable routes to theoretical understanding.

\underline{\it The ghost of Occam's razor: overfitting peaks and double descent} 

{\color{black} The discussion so far may provide the temptation to entirely abandon textbook wisdom and freely overparameterize and interpolate data. However, data interpolation by itself provides no guarantee of good generalization. One can trivially produce bad data interpolating functions ({\it e.g.} one that gives the training values on the training set, but is zero elsewhere). More interestingly, as one increases the numbers of fitting parameters during a parametric fit, the generalization error can {\it diverge} precisely when data interpolation is achieved, followed by a region of lower generalization error where regularized interpolation is performed using more parameters than data points. 

This phenomenon was first reported over 30 years ago (\cite{loog2020brief}, also see Figs 4.3 and 5.3 in \cite{engel2001statistical}) but was an obscure topic until recently. It has recently been highlighted and dubbed "double descent" \cite{belkin2019reconciling} and is an active research topic\cite{hastie2019surprises, montanari2019generalization, mitra2019understanding, adlam2020neural}. The diverging generalization error at the data interpolation point can be simply understood by examining multivariate linear regression with random design matrices. Consider the linear regression problem given by the data fitting model $Y=X\beta$, where $Y$ is an $m-$dimensional vector of observations, fitted by a $p$ dimensional parameter vector $\beta$, given the $m\times p$ matrix of regressors $X$. Precisely at the data interpolation point $m=p$, $X$ is square and will generically have singular values that tend to zero as the matrix becomes large. Since the least square parameter estimator is  $\hat{\beta}=(X^{\dagger}X)^{-1}XY$, ill-posedness of $(X^{\dagger}X)$ causes $\hat{\beta}$ and therefore the generalization error to diverge

With over-parameterization ($p>m$), the design matrix is rectangular and no longer has close to zero singular values as long as the null space is suppressed through appropriate regularization ({\it e.g. } by employing a minimum norm condition). The interesting point is that under suitable conditions, the generalization error in the over-parameterized region $p>m$ can in fact be lower than the GE in the under-parameterized region $p<m$. An intuitive scenario for this phenomenon, is the case where the number parameters $n$ in the data generating model is larger than $m$, and good generalization requires $p\geq n>m$ \cite{mitra2019understanding, hastie2019surprises}. In this case, having fewer parameters than data points $p<m$ causes model mis-specification, and this can cause the GE to suffer in the under-parameterized regime. Note however that these interpolating solutions are generally {\it not statistically consistent} and have generalization error above the theoretical lower limits. Another point to be noted, is that training data interpolation, in the sense of zero training error, can be achieved for classification, without passing through a peak in the GE curve. This is the case for boosting \cite{schapire1998boosting}. 

The non-monotonic behavior of the GE curves with increasing model complexity (the double descent scenario) is of interest in reconciling the textbook wisdom about the dangers of "overfitting" training data with the observations of good generalization under overparameterized conditions. However,the study of generalization error as a function of model complexity is not the focus of this perspective piece. Generally speaking, data interpolation is neither necessary nor sufficient for good generalization. The interpolating models in the overparameterized region of the double descent curves are not in general statistically consistent - the generalization error is above the theoretical lower bounds. While one can imagine that under certain circumstances data interpolation may produce reasonable generalization, what is surprising is the existence of interpolating learners that are {\it optimal} in the sense of Statistical Consistency. The provable examples available so far of such SCI learners (wiNN, Hilbert interpolation) are not parametric. }

{\bf SCI shows why ANNs in ML are distinct from  biological brains} 

\textcolor{black}{Artificial neural network models, particularly deep nets, have produced good results in practical applications. It is tempting to conclude (and this conclusion has been widely drawn) that the source of this success is that the ANNS/DNs mimic some important aspect of biological brains. What if this is not true? We argue in this article is that the success of ANNs/DNs in modern machine learning applications is the phenomenon of SCI, not necesssarily resemblance to biological brains. This has important implications.}

\textcolor{black}{First, although successful ANN/DN models for ML/RL were motivated loosely by real brain circuits, if they are indeed overparameterized data interpolation schemes owing their success to SCI, they are unlikely to provide a route to the scientific understanding of brains. Second, this holds open the possibility of a new generation of ML approaches that do indeed resemble real brain circuitry.}

\textcolor{black}{We first address the relation between real brains and ANNs. To the extent that ANNs are over-parameterized data interpolation devices whose generalization success is contingent on large training data sets - as this article would argue, and empirical evidence indicates - one is forced to conclude that they are quite different from biological brains. In contrast with the data-hungry interpolation algorithms, biological brains often need much smaller training data sets. For example, humans need to play many fewer games than the deep reinforcement learning algorithms to gain similar performance in video games\cite{hessel2018rainbow}. Striking examples of pre-programmed behaviors or "instincts" abound in the animal kingdom, where there is little or no requirement of training data. Even learned song culture in songbirds can be traced back to innate sources and can emerge spontaneously over multiple generations\cite{feher2009novo}. An obvious hypothesis is that biological brains are more efficient learners than modern ML algorithms due to extra "prior structure" (this is essentially the well known {\it poverty of stimulus} argument\cite{chomsky1988language}).} 

\textcolor{black}{Note that deep nets do have their own prior (handcrafted) network architecture ({\it e.g.} the encoder-decoder architecture and the translational invariance of convolutional layers in U-Net\cite{ronneberger2015u} like architectures for machine vision). Indeed such prior structure is essential for good performance even if deep nets are interpolating learners, and the "art" of designing deep nets consists in designing this structure in some manner. Nevertheless, the networks still require very large training data sets for good performance, and the training error is essentially driven to zero, indicating data interpolation. This is costly both from an engineering and financial perspective. Data interpolation implies sensitive dependence on the training data set and non-extrapolation, forcing the user to constantly gather new volumes of data as the application domain shifts. Adversarial examples are an obligatory property of interpolating learners, another engineering weakness. Also, overparameterization gives rise to very large models that are costly from an implementation perspective.}

\textcolor{black}{Thus, it would be beneficial for modern ML techniques to lessen the dependence on large volumes of training data by building in better priors. Since biological brains require smaller training data sets, it is natural to look at brains for inspiration for the necessary priors. How should this be done?}

{\bf Mesoscale brain circuit architecture as a route to incorporating biological priors into modern ML} 

\textcolor{black}{While ANNs bear a loose resemblance to biological neural networks, it is interesting to note that most of the biomimetic efforts so far have focussed on a behavioral/psychological level, drawing on reinforcement learning behaviors\cite{sutton2018reinforcement}, seeking to incorporate psychological/cognitive phenomena such as short term memory\cite{hochreiter1997long} or selective attention\cite{vaswani2017attention}. Also notable in this regard, is that Turing's inspiration for his model of computation came from the human cognitive behavioral process involved in solving mathematical problems, and the development of the formal theory of computer languages and automata relied on the analysis of linguistic phenomena. In all cases, brain-related considerations have largely occurred at the psycho-behavioral level rather than at the level of the underlying brain circuitry. However, despite these sources of inspiration for brain mimicry, modern ML appears to be still producing data-hungry interpolation schemes.}

\textcolor{black}{Is there a different approach to incorporating prior knowledge that will move us away from data interpolation schemes? We wish to argue that there is a currently under-exploited avenue that would be fruitful to explore in this context, {\it i.e.} "network mimesis", where the biomimetic effort is directed to the actual network architecture of real brains rather than surface behaviors or cognitive psychological constructs. Arguably, mimicry at the neural circuit level is the origin of ANNs, reflected in the work of McCullogh and Pitts\cite{abraham2002physio}. Turing also proposed an ANN model of computation inspired by neural circuitry\cite{turing1948intelligent}.}

\textcolor{black}{Deep nets for machine vision are loosely inspired by the hierarchical organization of the primate visual system\cite{lecun2015deep}. However, networks used in contemporary ML algorithms bear little resemblance to real neural circuit architecture - in the case of machine vision, it is routine to use dozens if not hundreds of feed forward layers, whereas the visual system employs perhaps five or six hierarchical levels, with strong feedback connections between levels\cite{sillito2006always}. As another example, the convolutional structure (equivalently, translationally invariant filters) used in CNNs for machine vision are in fact different from the real visual systems, which have 'receptive fields', ie different sets of neurons perform the spatial filtering operations at different points on the image, rather than a shared set of filters performing the operation across the entire network ("weight sharing").}

\textcolor{black}{Human engineers devising machine learning networks design their "network anatomy", so circuit architectures can be adopted in a straightforward manner. At what level should one try to adopt the circuit connectivity patterns from brains? We would like to suggest that a fruitful future avenue for neuromimesis lies in the mesoscale circuit architecture\cite{bohland2009proposal,mitra2014circuit,oh2014mesoscale} of entire vertebrate brains. Network architecture at the mesoscopic scale is concerned with the connectivity rules or patterns between cytoarchitectonically distinct brain compartments or groups of similar neurons. This mesoscale circuit architecture is species-specific and genetically encoded. In contrast, individual synaptic connections (as measured using Electron Microscopy, {\it e.g.} the recently published synaptic connectivity matrix of a major portion of the fruitfly brain\cite{scheffer2020connectome}) are individual-specific and subject to plasticity and learning. Indeed, one might argue that human engineers designing deep nets are precisely engaged in the engineering of mesoscale circuitry of ANNs, leaving the specification of individual "synaptic weights" to the process of fitting training data sets. }

As an example of how this might work, consider the recent neuroanatomical finding of unidirectional monosynaptic connections between the primary auditory cortex and the periphery of the primary visual cortex in the marmoset\cite{majka2019unidirectional}. This early fusion of the sensory streams stands in distinction to how video is processed today, where the image and sound streams are first converted into high level symbols before fusion; the mesoscale projection from primary auditory to primary visual cortex might argue for the utility of early fusion of the data streams. 

\textcolor{black}{Beyond this specific example, there are many prominent differences between deep net architectures and mesoscale brain architectures. The mesoscale network architecture in real brains is "shallow-recurrent"\cite{sillito2006always} rather than "deep-feedforward". In the visual pathway, often used to motivate deep networks, there are only five or six levels before object recognition is achieved, and strong feedback connections exist from early on (cf. the strong projection from the primary Visual cortex to the sensory relay neurons in the Lateral Geniculate Nucleus of the Thalamus). One field where such input from mesoscale circuitry may prove particularly useful is robotics - this is an area where deep nets have had limited impact\cite{kaelbling2020foundation}. The circuit architectures of entire vertebrate brains, particularly the integration between sensory and motor systems and the hierarchical organization of motor systems may prove of utility ({\it c.f.} recent work elaborating the input-output projections of the mouse motor cortex\cite{munoz2020cellular}).} 

\textcolor{black}{Behavioral/psychological priors tell us more about the problem being solved than about the algorithm being employed to solve the problems. Neuroanatomical circuitry {\it are} in some sense the distributed algorithms that are involved in the problem solution and closer examination of these circuits may help close the training data gap between data-interpolating deep nets and biological brains. }

{\bf Understanding Nature: Scientific Theories vs Data Interpolation}

\textcolor{black}{A second set of questions raised by data interpolation in modern ML relates to scientific models and theories. As the quote from Von Neumann shows, models or theories with too many parameters are generally regarded as problematic in scientific work, and interpolating noisy data is antithetical to basic practice in science. However, to the extent that scientific theories are meant to be predictive, statistically consistent interpolation methods also offer predictions that are optimal in the limit of many observations. Can they therefore replace scientific models and theories? Such a replacement would have big implications for the training of scientists and for scientific methodology: learning and developing theories in the classical manner would simply be replaced by large scale data gathering exercises\cite{schmidt2009distilling}. While such an enterprise would be centuries too late in terms of discovering physical laws, it raises provocative questions about future scientific analysis of complex phenomena. }

\begin{figure}
  \centering
    \includegraphics[width=\linewidth]{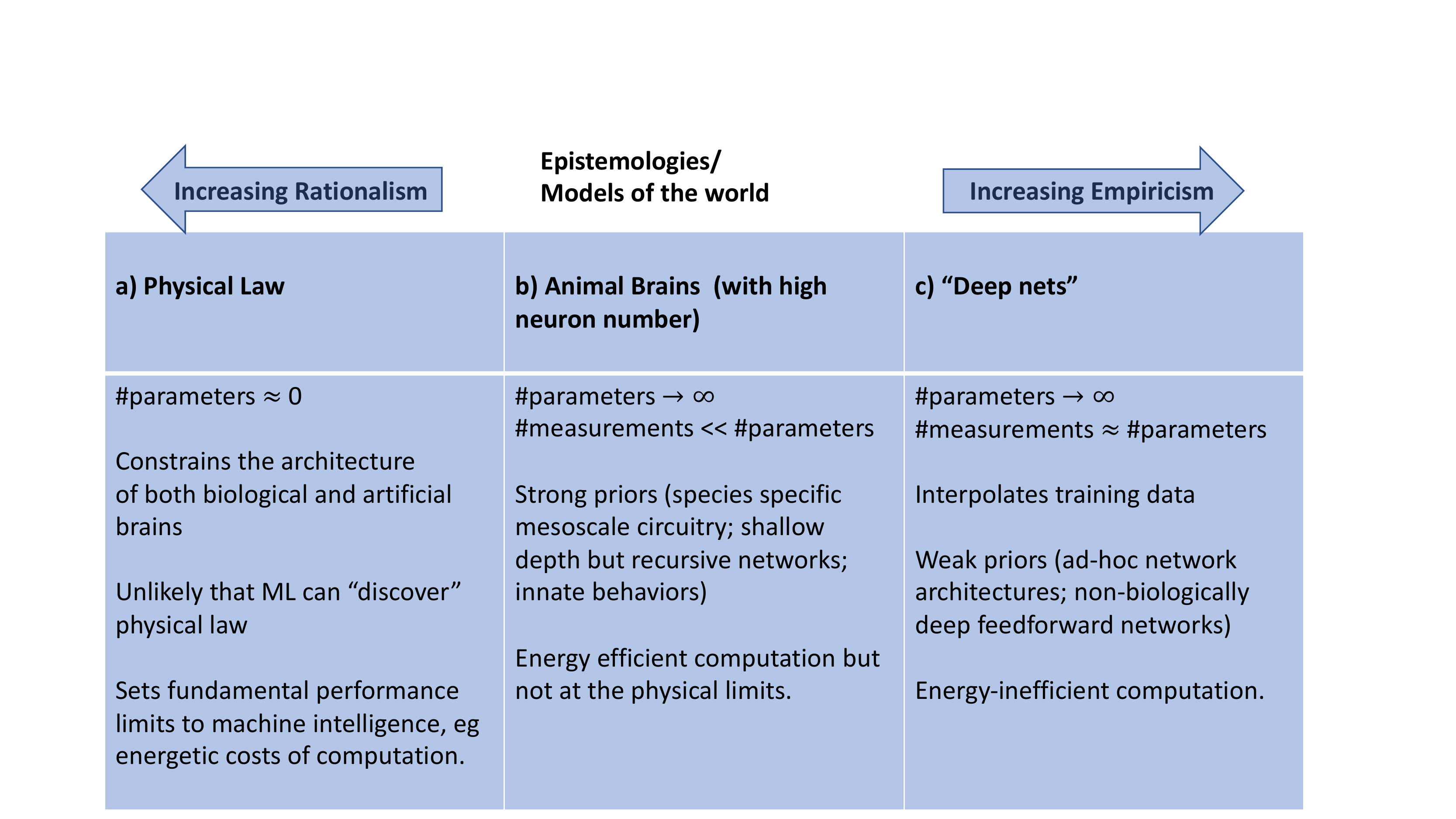}
\caption{The figure summarizes the perspective presented in this article relating the three epistemologies or theories of the world as viewed by the laws of physics (a), by animal brains with large neuron numbers (b) and by deep nets (c). Physical law is strongly rationalist in spirit, adopting {\it a priori} theories with few if any adjustable parameters that are revised when contradicting empirical evidence. Complex animal brains with large neuron numbers have exponentially many parameters (potential synaptic connectivity), but have innate structure and efficient learning mechanisms that require a modest set of empirical measurements. These innate predispositions of complex animal brains are captured by the mesoscale circuit architectures specified by the species genomes. Modern ML as exemplified by deep net architectures is strongly empiricist, with a large number of parameters that are fitted to an equally large number of measurements, often in an interpolating manner, with relatively weak priors which are imposed by relatively ad-hoc, engineered network architectures. Deriving physical law by fitting data in the spirit of modern ML is inconsistent with previously successful approaches. Physical law sets fundamental energetic limits to both biological brains and current implementations of ML algorithms, and the current hardware implementations of ML appear to be highly energetically inefficient, operating far from the fundamental physical limits.}
\end{figure}

\textcolor{black}{SCI learners perfectly "explain" (fit) the observations while generalizing well - superficially this is also what one expects of scientific theories. An obvious weakness is the sensitive dependence of interpolating learners on the distribution from which the training set is chosen. Non-stationarity of the training distribution and the appearance of previously unobserved regimes in the test will prevent interpolating approaches from predicting accurately outside the training data domain. However, one might consider a brute force approach that continuously gathers more training data as distributional non-stationarities are encountered, in real life engineering applications. This may well succeed in narrow domains of application. }

\textcolor{black}{There is however a basic difficulty in trying to automate scientific knowledge generation by interpolating data. Indeed, the success of modern physics, a paradigmatic science, is not based on trying to explain the full complexity of observations}, a strategy that Noam Chomsky has characterized as predicting "what's gonna happen outside the window next"\cite{katz2012noam}. Interpolating learners may indeed be better tools for explaining the intricacies of the world outside the window ({\it if} equipped with suitably large training corpora), but the successful development of theoretical physics depended on ignoring the world outside the window, so to speak, and looking into the night sky to study planetary orbits. \textcolor{black}{Interestingly, the first models of planetary orbits in fact resemble the modern Machine Learning approach using function fits: the method of epicycles used in Ptolemy's astronomy to predict planetary orbits, are in fact equivalent to a Fourier series expansion\cite{acosta2020need}.}

Thus a key to theoretical advancements in physics has been to {\it ignore} the observed complexities of the natural world and to selectively focus on well-chosen details that provide the right hooks to unravel the seeming complexity. This strategy has clearly been very successful. Overparametrized and interpolating learners {\it by design} do not provide a methodology for selecting a few "important" details. Therefore, despite pragmatic success in solving real problems they seem unsuitable for scientific theorizing. 

\textcolor{black}{The perspective presented in this article may be summarized as follows. The success of data-intensive modern machine learning approaches depend in an important way on the phenomenon of Statistically Consistent Interpolation. SCI learners generalize well while interpolating training data in the limit of large data sets, a property that seemingly contradicts previous statistical practice, but that can be proven rigorously for some data interpolation algorithms.  To the extent that modern ML methods are interpolation schemes, implemented using overparametrized models dependent on very large data sets for good generalization, they present a {\it third} way of modeling or predicting natural phenomena, in contrast with theoretical physics, which has very few parameters, but also in contrast with biological brains, which effectively have many parameters but do not require such large training data sets. This points to the possibility of a new generation of intelligent machinery based on distributed circuit architectures which incorporate stronger priors, possibly drawing upon the mesoscale circuit architecture of vertebrate brains. }

\textcolor{black}{One important theme that is beyond the scope of this perspective piece, but which bears mention, is the matter of energetic costs. Implementations of modern machine learning on existing hardware approaches are energy inefficient compared to biological brains. In addition to the "data gap" discussed in this article, this "energy gap" is an important area of future study. Physical law may set fundamental limits to the energetic efficiency of intelligent machines\cite{landauer1961irreversibility}, and it is also possible that non-biological hardware and computational paradigms may permit yet other varieties of machine intelligence we have not yet conceived\cite{feynman2018feynman}.}

\textcolor{black}{Given the fundamental weakness of data interpolation as a repository of observations, it is unlikely that theoretical physicists will switch to fitting elephants. Neither will overparametrized, data-fitting ANNs help us understand the mystery of how and why our biological brains can imagine perfectly equal Platonic sticks without any experiential basis. Nevertheless, we now have a better theoretical understanding of how one can interpolate noisy data and generalize well at the same time, an observation that is likely to have lasting theoretical as well as practical impact across multiple disciplines. }

\section*{Acknowledgement} This work was supported by the Crick-Clay Professorship (CSHL) and the H N Mahabala Chair Professorship (IIT Madras). 
\pagebreak

\bibliography{main}
\bibliographystyle{unsrt}

\end{document}